\documentclass[journal]{IEEEtran}
\usepackage{fancyhdr}

  \usepackage{amsmath}
  \usepackage{amssymb}
  \usepackage[nocompress]{cite}
  \usepackage{graphicx}
  \usepackage{multicol}
  \usepackage{ragged2e}
  \usepackage{booktabs}
  \usepackage{multirow}
  \usepackage[table,xcdraw]{xcolor}
  \usepackage[normalem]{ulem}
  \usepackage{epsfig}
  \usepackage{subcaption}
  \usepackage{caption}
  \usepackage{amsfonts}
  \usepackage{adjustbox}
  \usepackage{bbding}
  \usepackage{makecell}
  \usepackage{arydshln}
  \usepackage{colortbl}
  \usepackage{overpic}
\usepackage[pagebackref=false,
            breaklinks=true,colorlinks,
            citecolor=cyan,linkcolor=red,
            bookmarks=false]{hyperref}

\begin{document}

\title{Traffic Image Restoration under Adverse Weather via Frequency-Aware Mamba}

\author{Liwen Pan,
        Longguang Wang,
        Guangwei Gao,
        Jun Wang,
        Jun Shi,
        Juncheng Li\textsuperscript{$\ast$}

\thanks{Liwen Pan, Jun Wang, Jun Shi are with the School of Communication and Information Engineering, Shanghai University, Shanghai, China.}% <-this % stops a space
\thanks{Longguang Wang is with the College of Electronic Science, Air Force Aviation University, 130022, China.}% <-this % stops a space
\thanks{Guangwei Gao is with the Intelligent Visual Information Perception Laboratory, Institute of Advanced Technology, Nanjing University of Posts and Telecommunications, Nanjing, China.}
\thanks{Juncheng Li is with the School of Computer Science and Technology, East China Normal University, Shanghai, China (e-mail: jcli@cs.ecnu.edu.cn).}
}

%\iffalse
\markboth{IEEE}%
{Shell \MakeLowercase{\textit{et al.}}: Bare Demo of IEEEtran.cls for IEEE Journals}
%\fi

\maketitle

\begin{abstract}
Traffic image restoration under adverse weather conditions remains a critical challenge for intelligent transportation systems. Existing methods primarily focus on spatial-domain modeling but neglect frequency-domain priors. Although the emerging Mamba architecture excels at long-range dependency modeling through patch-wise correlation analysis, its potential for frequency-domain feature extraction remains unexplored. To address this, we propose Frequency-Aware Mamba (FA-Mamba), a novel framework that integrates frequency guidance with sequence modeling for efficient image restoration. Our architecture consists of two key components: (1) a Dual-Branch Feature Extraction Block (DFEB) that enhances local-global interaction via bidirectional 2D frequency-adaptive scanning, dynamically adjusting traversal paths based on sub-band texture distributions; and (2) a Prior-Guided Block (PGB) that refines texture details through wavelet-based high-frequency residual learning, enabling high-quality image reconstruction with precise details. Meanwhile, we design a novel Adaptive Frequency Scanning Mechanism (AFSM) for the Mamba architecture, which enables the Mamba to achieve frequency-domain scanning across distinct subgraphs, thereby fully leveraging the texture distribution characteristics inherent in subgraph structures. Extensive experiments demonstrate the efficiency and effectiveness of FA-Mamba.
\end{abstract}


\section{Introduction}
One of the great challenges facing intelligent transportation systems is how to work effectively in adverse weather conditions. However, traffic images captured under such conditions often suffer from visual degradation, significantly affecting downstream visual tasks such as semantic segmentation, object detection, and autonomous driving~\cite{Zhao_2017_CVPR, wu2019detectron2,Hu_2023_CVPR,10024313,10323218}. To overcome this challenge, it is crucial to restore clean images from degraded ones, such as rain or snow images. However, it is a pretty difficult task due to occlusion and the loss of important perceptual frequency information in the images.

\begin{figure}[htbp]
\centering
\includegraphics[width=0.5\textwidth]{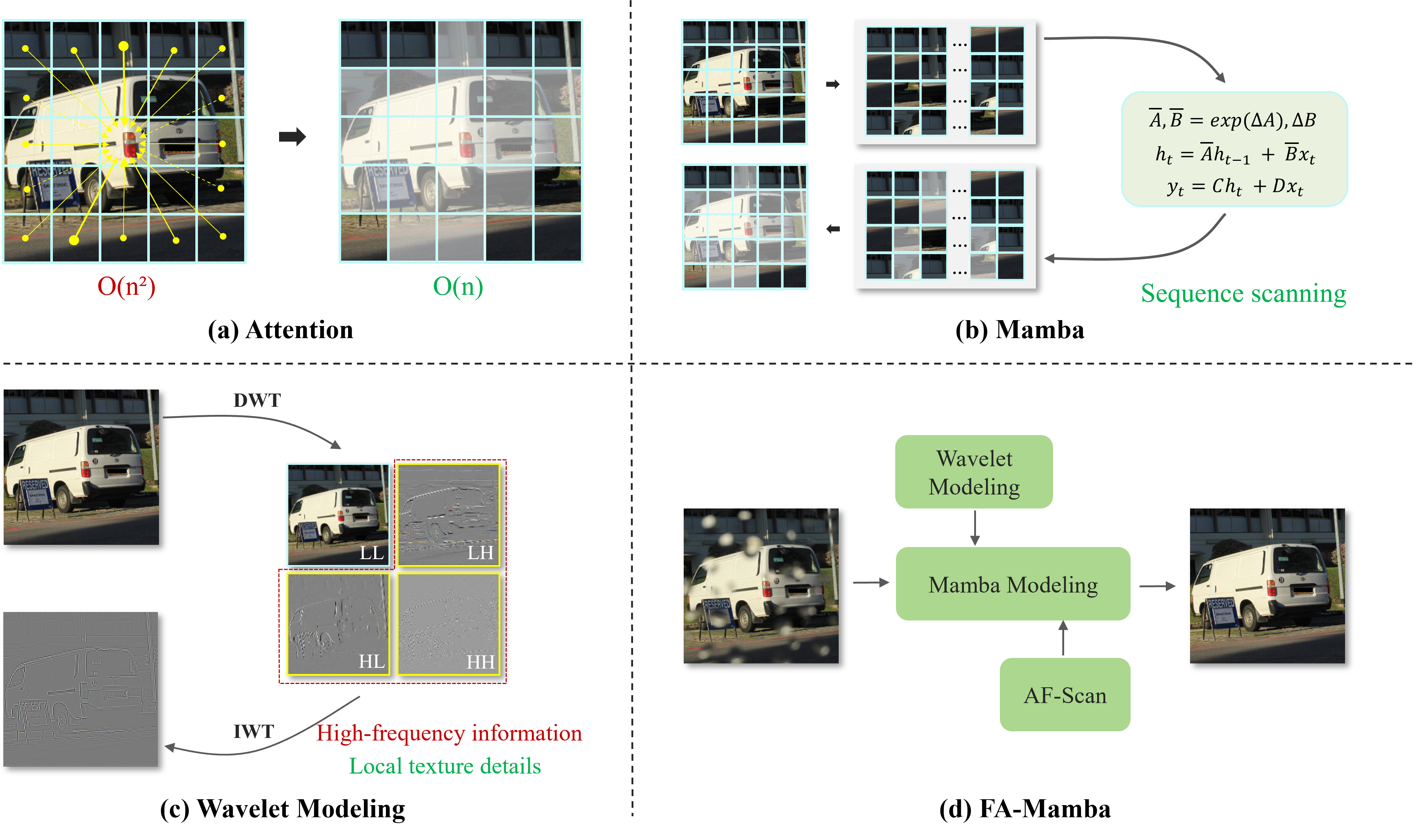}
\caption{Comparison of different modeling methods. Our FA-Mamba can enhance the ability of the Mamba to restore image texture details from the frequency perspective.}
\label{fig:comparison of different modeling methods}
\end{figure}

Most traditional methods rely on prior knowledge or estimate task-specific degradation functions to achieve image restoration, but these methods usually have mediocre results and poor robustness. Recently, deep learning-based methods~\cite{Fu_2017_CVPR,Jiang_2020_CVPR,liang2021swinir,zamir2022restormer,wang2022uformer,wang2024gridformer,10359459} have revolutionized image restoration by utilizing various deep network architectures, such as CNN and Transformer, to learn the mapping relationship between images and obtain higher quality images than traditional methods. Especially, the attention mechanism in Transformers has achieved remarkable success due to its ability to capture and model internal correlations within an image. While, as illustrated in Fig.~\ref{fig:comparison of different modeling methods}, the attention mechanism encounters scalability issues because of its quadratic complexity, which becomes a major challenge when processing large images. To address these limitations, the improved structured state-space sequence model (S4) with a selective scanning mechanism, Mamba, stands out for its ability to model long-range sequence dependencies with linear complexity. However, it is important to note that Mamba relies solely on pixel sequences to model image degradation in visual tasks, which limits its ability to perceive degradation in both spatial and frequency domains. To solve this issue, we aim to extend it to the wavelet domain and restore the image from both spatial and frequency domains.

As is well known, the low-frequency information in the wavelet transform mainly corresponds to the global structure of the image, while the high-frequency information primarily captures details and textures. In image restoration tasks, recovering texture details is both crucial and challenging. To address this,  we leverage the ability of Mamba to capture regional correlations in the spatial domain while integrating high-frequency analysis to perceive degradation better. This enhances the capability of the model to restore intricate textures more effectively. 

Specifically, we develop an effective Frequency-Aware Mamba for image restoration, named FA-Mamba. The infrastructure of FA-Mamba consists of a series of FA-Blocks, which include two key components:  the Dual-Branch Feature Extraction Block (DFEB) and the Prior-Guided Block (PGB). The DFEB (Dual-Branch Feature Extraction Block) represents a hybrid module that integrates Mamba and CNN architectures, enabling the extraction and fusion of both local and global features. Compared to Transformer-based models, DFEB significantly reduces computational resource consumption while maintaining competitive performance. Meanwhile, the PGB (Prior-Guided Block) introduces enhanced high-frequency information representation, thereby improving the capability of the model in capturing intricate image texture details. Furthermore, we design a novel Adaptive Frequency Scanning Mechanism (AFSM) for the Mamba architecture. This mechanism facilitates frequency-domain scanning across distinct subgraphs, thereby fully leveraging the texture distribution characteristics inherent in subgraph structures. By introducing this innovative scanning approach from the frequency-domain perspective, AFSM offers a new method for exploration within the two-dimensional spatial domain.

Our contributions can be summarized as follows:
\begin{itemize}
\item We propose an efficient Frequency-Aware Block (FA-Block), which is composed of a Dual-Branch Feature Extraction Block (DFEB) and a Prior-Guided Block (PGB). This module is capable of simultaneously capturing both local and global features of the image and leveraging high-frequency information to generate high-quality restoration results.
\item We design a novel Adaptive Frequency Scanning Mechanism (AFSM) for the Mamba architecture. This mechanism facilitates frequency-domain scanning across distinct subgraphs, thereby fully leveraging the texture distribution characteristics inherent in subgraph structures.
\item We propose a Frequency-Aware Mamba (FA-Mamba) for traffic image restoration under adverse weather. Extensive experiments demonstrate the efficiency and effectiveness of FA-Mamba.
\end{itemize}

\section{Related Works}
\subsection{Image Restoration}
The field of image restoration under adverse weather conditions has evolved significantly, transitioning from early model-based methodologies to sophisticated data-driven approaches. Traditional model-based techniques~\cite{he2010single, luo2015removing, zhu2017joint, 9794328} primarily rely on the identification of appropriate weather-specific priors to address the image restoration problem. With the advent of deep learning in recent years, image restoration has achieved remarkable advancements in reconstruction quality. Foundational works such as~SRCNN \cite{dong2015image} and DnCNN~\cite{zhang2017beyond} leveraged compact convolutional neural networks (CNNs) to deliver impressive performance in image restoration tasks. Nevertheless, CNN-based methods typically encounter challenges in effectively modeling global dependencies across image structures.

Subsequently, the Transformer architecture—originally developed for natural language processing—has demonstrated extraordinary potential in computer vision and image restoration, rapidly emerging as a cornerstone for such tasks~\cite{liang2021swinir, wang2022uformer, zamir2022restormer, wang2024gridformer}. The Transformer excels in capturing global dependencies and modeling intricate relationships, which CNNs struggle to address. However, its self-attention mechanism introduces computational complexity that scales quadratically with input image size, posing significant challenges for larger image restoration tasks where computational efficiency becomes increasingly critical. Recently, diffusion models have garnered substantial attention in image restoration~\cite{cheng2024weafu, 10420512, 10021824}, primarily due to their exceptional generative capabilities that enable precise data distribution modeling. Despite these advantages, the high computational resource requirements and prolonged inference times remain significant barriers to their widespread adoption in practical applications.

\subsection{State Space Models}
Originating from classical control theory~\cite{10.1115/1.3662552}, State Space Models (SSMs)~\cite{gu2021efficiently, gu2021combining, smith2022simplified} have attracted substantial attention from researchers due to their unique capability to process long-range dependencies with linear scalability relative to sequence length, thereby achieving remarkable computational efficiency in deep learning applications.

A particularly noteworthy advancement is Mamba~\cite{gu2023mamba}, a selective, data-aware SSM specifically optimized for hardware acceleration. Mamba has demonstrated superior performance to Transformers in natural language processing tasks while maintaining linear scaling with input length, which is an essential advantage for large-scale data. Beyond NLP, initial applications of Mamba to vision tasks have shown promise, including image classification~\cite{liu2024vmamba}, image restoration~\cite{guo2024mambair}, biomedical image segmentation~\cite{ma2024u}, and other areas~\cite{hu2024zigma, zha2024lcm}. This study aims to explore the potential of Mamba for image inpainting and develop a simple yet effective traffic image restoration framework based on this architecture.

\subsection{Wavelet Transform-Based Methods}
Wavelet transform has become a widely adopted technique in image processing tasks. With the advent of deep learning, researchers have explored its integration with CNNs and Transformers, yielding promising results. For instance, Liu et al.~\cite{Liu_2018_CVPR_Workshops} proposed a Multi-level Wavelet Convolutional Neural Network (MWCNN) for image restoration, effectively leveraging multi-scale wavelet decomposition to enhance processing. Similarly, Li et al.~\cite{LI2024106378} introduced the Efficient Wavelet Transformer (EWT), which accelerates the computational speed of traditional Transformers and substantially reduces GPU memory consumption, all while maintaining superior restoration quality. In the realm of generative models, Huang et al.~\cite{10420512} developed the Wavelet-Based Diffusion Model (WaveDM), which models clean image distributions directly in the wavelet domain to address the inherent issue of prolonged inference times. Inspired by these advancements, we aim to investigate the efficacy of Mamba within the wavelet domain and propose a lightweight Mamba variant leveraging wavelet transform techniques.

\begin{figure*}[ht]
\centering
\includegraphics[width=1\textwidth]{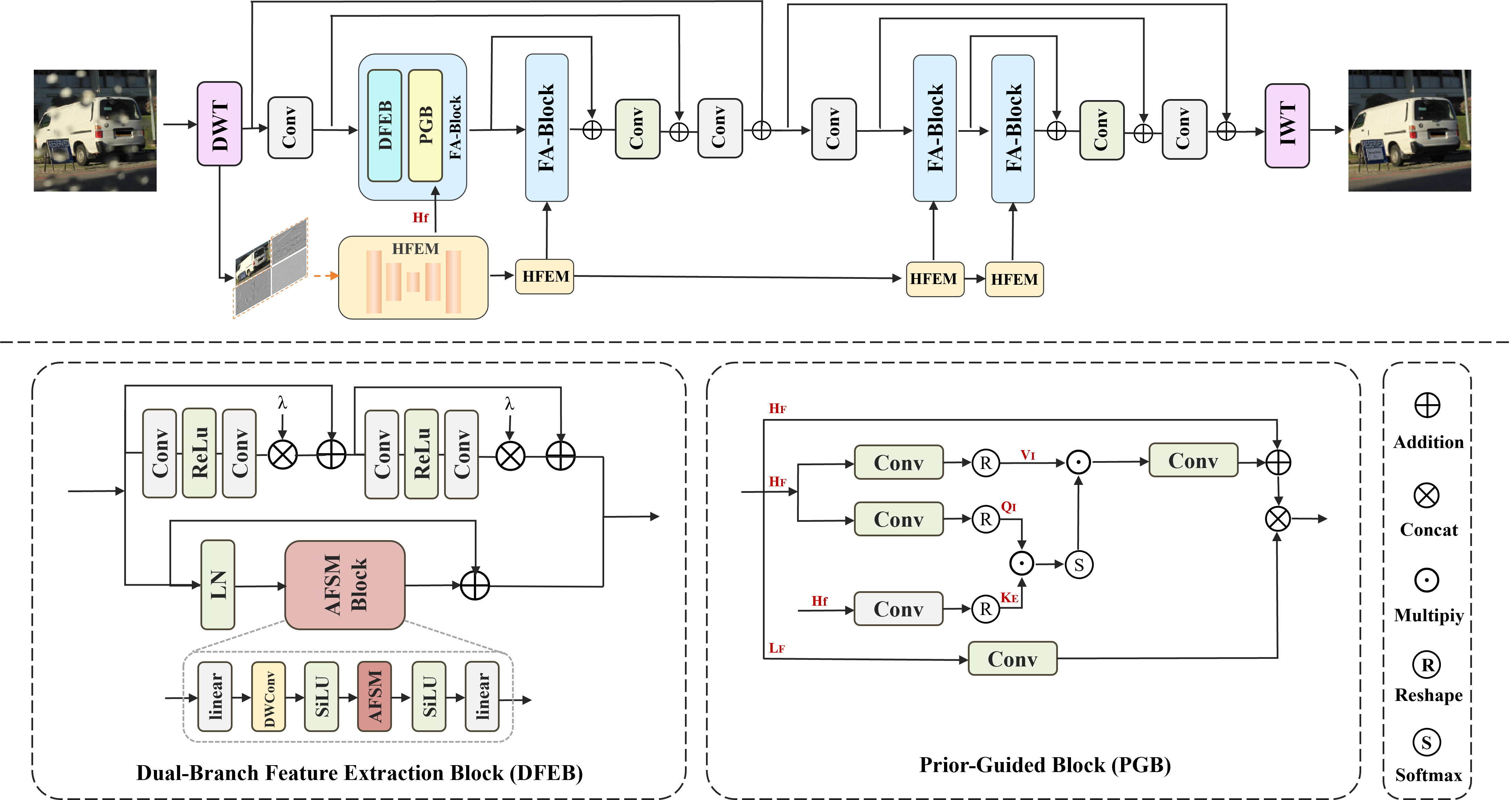}
\caption{The overall architecture of the proposed Frequency-Aware Mamba (FA-Mamba). FA-Mamba mainly contains a series of Frequency-Aware Blocks (FA-Blocks) and High-Frequency Enhancement Module (HFEM). Specifically, the FA-Block is composed of a Dual-Branch Feature Extraction Block (DFEB) and a Prior-Guided Block (PGB).}
\label{fig:overall architecture}
\end{figure*}

\section{Method}
% In this section, we first introduce the preliminaries of frequency analysis and SSMs. Afterward, we outline the overall framework of FA-Mamba. Then we provide the details of the proposed frequency-aware block (FAB) as the fundamental building unit of our method, which mainly contains two key elements:  Dual-Branch feature extraction block (DFEB) and a Prior-Guided block (PGB). Finally, we present the adaptive frequency scanning mechanism (AFSM).
\subsection{Preliminaries}\label{AA}
\subsubsection{\textbf{Frequency Analysis in Digital Imaging}}
The Wavelet Transform is a powerful mathematical tool that decomposes an image $I_{in} \in \mathbb{R}^{H \times W \times C}$ into four sub-bands representing low-frequency approximations and high-frequency details, capturing both spatial and frequency information simultaneously. The low-frequency components preserve the overall contour information, while the high-frequency components capture details such as edges and textures.  The Discrete Wavelet Transform (DWT), a discrete-time implementation of wavelet analysis, mathematically operates cen be defined as:
\begin{equation}
I_{LL}, I_{LH}, I_{HL}, I_{HH} = f_{\text{DWT}}(I_{in}),
\end{equation}
where $I_{LL}, I_{LH}, I_{HL}, I_{HH} \in \frac{H}{2} \times \frac{W}{2} \times C$
are 4 sub-images with different frequencies. With the help of DWT, high-frequency and low-frequency features can be effectively separated while reducing the image resolution.

\subsubsection{\textbf{State Space Models}}
State Space Models (SSMs) serve as a fundamental framework for mapping one-dimensional inputs to outputs through latent states, leveraging linear ordinary differential equations. Given a system with input \( x(t) \) and output \( y(t) \), its dynamics can be expressed as follows:
\begin{equation}
\begin{aligned}
\bar{A}, \bar{B} &= \exp(\Delta A), \Delta B, \\
h_t &= \bar{A} h_{t-1} + \bar{B} x_t, \\
y_t &= C h_t + D x_t.
\end{aligned}
\end{equation}

To enable the application of continuous-time systems in deep learning frameworks, the zero-order hold (ZOH) method is employed with a time-scaling parameter \(\Delta\), facilitating precise discretization. This technique effectively transforms the continuous system parameters \(A\) and \(B\) into their discrete equivalents \(\bar{A}\) and \(\bar{B}\), ensuring numerical stability while preserving system dynamics. Furthermore, the discretized formulation enables adaptive scanning mechanisms that dynamically adjust to input data characteristics. Such adaptability is crucial in complex tasks such as image restoration, where accurately capturing long-range contextual dependencies between image regions is paramount for high-quality reconstruction.

\subsection{Overall Architecture}
The architectural design of our proposed Frequency-Aware Mamba (FA-Mamba) is depicted in Fig.~\ref{fig:overall architecture}. Given a degraded image $I \in \mathbb{R}^{H \times W \times 3}$, we initially apply the Discrete Wavelet Transform (DWT) to obtain a multi-scale representation. Subsequently, convolutional operations are employed to extract shallow features $X_{f} \in \mathbb{R}^{H \times W \times 3}$. Concurrently, the High-Frequency Enhancement Module (HFEM) is utilized to refine the high-frequency components, yielding an enhanced prior map $X_{hf} \in \mathbb{R}^{H \times W \times 3}$. The extracted shallow features $X_{f}$ are then processed through a series of Frequency-Aware Blocks (FA-Block) to perform hierarchical deep feature extraction. This architecture not only significantly accelerates the inference speed but also ensures efficient utilization of high-frequency information during feature processing.  

\begin{figure*}[h]
\centering
\includegraphics[width=1\textwidth]{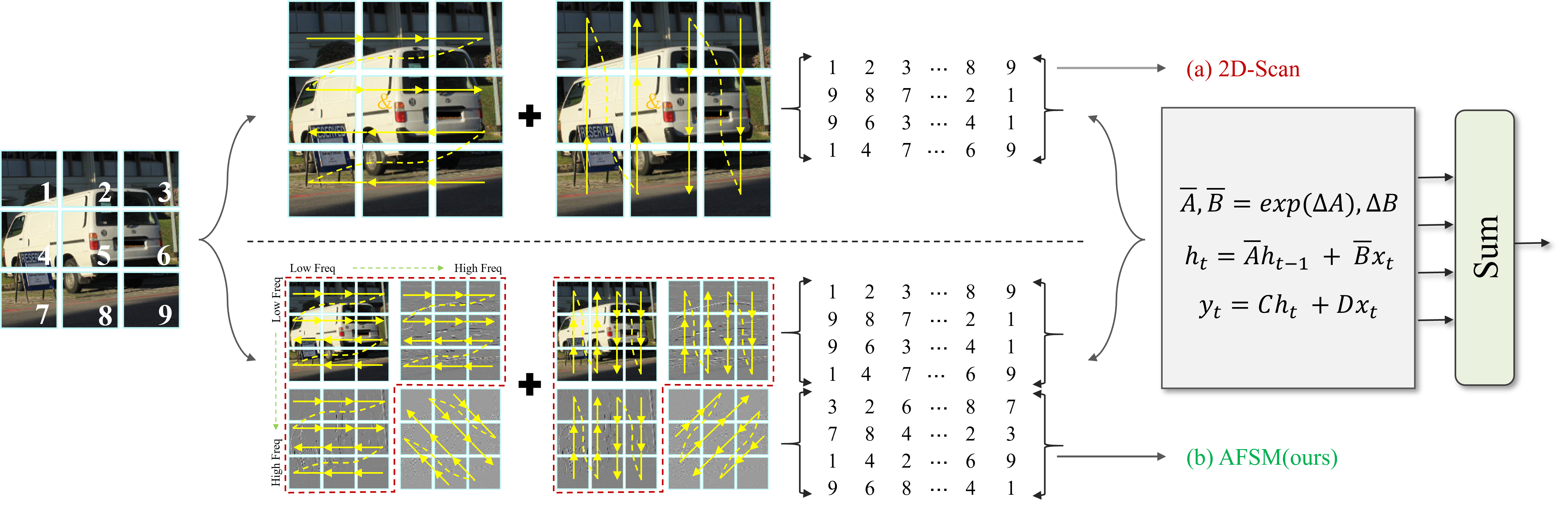}
\caption{The comparison between (a) the 2D scanning strategy employed by VMamba\cite{liu2024vmamba} and (b) our proposed Adaptive Frequency Scanning Mechanism (AFSM). Utilizing the wavelet Transform, we decompose the input into 4 frequency bands. It is then scanned along the frequency dimension in the spatial domain. Specifically, this strategy captures complex image details at different frequencies by relying on the texture distribution of subgraphs and applying a feature scanning mechanism.}
\label{fig:aptive Frequency
Scanning Mechan}
\end{figure*}

In each FA-Block, unlike the conventional self-attention mechanism, we employ the Mamba architecture for global modeling via efficient 2D sequential scanning. Subsequently, the Prior-Guided Block (PGB) is utilized to further refine the extracted features by enhancing local details and texture structures, guided by the high-frequency prior map $X^{0}_{hf} \in \mathbb{R}^{H \times W \times 3}$. To preserve hierarchical information, skip connections are employed to fuse hierarchical features effectively. Finally, the Inverse Wavelet Transform (IWT) is applied for upsampling, enabling accurate image restoration and the reconstruction of the corresponding clean image.  

\subsection{Frequency-Aware Block (FA-Block)}
FA-Block is the core component of FA-Mamba. The entire network contains 4 FA-Blocks, and each FA-Block consists of a Dual-Branch Feature Extraction Block (DFEB) and a Prior-Guided Block (PGB). As the most important module of the model, it can extract rich image features and utilize high-frequency information to generate high-quality restoration results. In the following sections, we will introduce DFEB and PGB in detail.

\subsubsection{\textbf{Dual-Branch Feature Extraction Block (DFEB)}}
A multi-branch architecture facilitates the extraction and integration of multi-scale information, thereby enhancing the representational capacity of the model. To leverage this advantage, we propose the Dual-Branch Feature Extraction Block (DFEB), which is specifically designed to capture complementary hierarchical features by employing parallel processing branches. Specifically, we replace traditional multi-receptive-field mechanisms with the Mamba module, forming a hybrid architecture comprising Mamba and CNN branches to extract distinct yet complementary image features. As illustrated in Fig.~\ref{fig:overall architecture}, DFEB integrates two specialized branches: one is used to extract global information and the other is used to capture local details. The dynamics are as follows:
\begin{equation}
\begin{aligned}
X_{e1} = f_{\text{CNN}}(X_{f}), \\
X_{e2} = f_{\text{Mamba}}(X_{f}), \\
X_{out} = X_{e1} + X_{e2}.
\end{aligned}
\end{equation}
where $X_{f}$ and $X_{out}$ are the input and output of the DFEB, respectively. Meanwhile, the CNN branch consists of several convolutional layers and ReLU activation functions. It is worth mentioning that the Mamba branch introduces our specially designed adaptive frequency scanning mechanism, which can better scan and extract features in the frequency domain.

\subsubsection{\textbf{Adaptive Frequency Scanning Mechanism (AFSM)}}
The 2D scanning strategy (Fig.\ref{fig:aptive Frequency Scanning Mechan} (a)) employed by VMamba exhibits several distinctive characteristics that enhance its efficiency and effectiveness in processing spatial features. Specifically, this approach enables sequential feature aggregation along predefined scanning paths, facilitating contextual dependency modeling and computational efficiency by reducing redundant operations. However, this scanning method does not apply to the frequency domain because the wavelet transform divides the image into four sub-images, each representing different frequency information. 

Inspired by the LocalMamba framework \cite{huang2025localmamba}, we design an Adaptive Frequency Scanning Mechanism (AFSM) to enhance feature extraction. As depicted in Fig.\ref{fig:aptive Frequency Scanning Mechan} (b), AFSM acts on four wavelet subbands after DWT processing and applies a customized scanning strategy to each subband. Specifically, low-frequency sub-bands (LL, LH, HL) employ horizontal and vertical scanning to capture low-frequency global features. On the contrary, diagonal scanning is specifically employed on the high-frequency sub-band (HH) to capture high-frequency features such as edges and corners by prioritizing short-range dependencies, which enables efficient extraction of fine-grained spatial details. With the help of this scanning method, we can achieve frequency information perception and cross-subband scanning, achieving a better balance between computational efficiency and information extraction.

\subsubsection{\textbf{Prior-Guided Block (PGB)}}
The core idea of this block is to leverage high-frequency priors for guiding the model to reconstruct clean images with precise texture details. Most existing methods typically employ simple concatenation or additive operations to integrate image features with texture information. However, these methods often inadequately exploit the available high-frequency priors. To address this limitation, we introduce a novel Prior-Guided Block (PGB) operating in the wavelet domain, which serves as an efficient attention mechanism that effectively utilizes enhanced high-frequency components to direct the model towards generating superior quality reconstructions with significantly improved detail accuracy.

As shown in Fig.~\ref{fig:overall architecture}, PGB takes the output $X_{out}$ of DFEB and the output $X_{hf}$ of High-Frequency Enhancement Module (HFEM) as inputs. For $X_{out}$, we extract its high frequency components $X^{HF}_{out}$ and generate the corresponding image queries $Q_{I}$ and image values $V_{I}$ by sequentially applying 1×1 convolutions and 3×3 depth-wise convolutions to encode the context. Similarly, we generate the pior keys $K_{E}$ by applying convolutions to project the enhanced high-frequency information $X_{hf}$ into the same dimension space as the queries. Specifically, HFEM recives the high-frequency sub-bands (LH, HL, HH) obtained from the wavelet decomposition of the input image with a U-net architecture for both encoding and decoding these components, effectively preserving the most critical feature information. It is worth noting that HFEM is weight-shared module in the model. After the above operations, $Q_{I}$, $K_{E}$, and $V_{I}$ have the same shape as $\frac{H}{2} \times \frac{W}{2} \times 3C$. Then, we reshape $Q_{I}$ and $K_{E}$ to $3C \times\frac{HW}{4}$, and generate the channel attention weights between $Q_I$ and $K_E$ through their dot-product interaction. In this manner, high-frequency information can further enhance image features and help reconstruct higher-quality images. The overall PGB process can be formulated as follows:
\begin{equation}
\begin{aligned}
X_{\text{res}} = V_I \cdot \text{Softmax}(K_E \cdot Q_I).
\end{aligned}
\end{equation}
Finally, we add \(X_{\text{res}}\) and \(X^{Hf}_{out}\), then concatenate the result with the low-frequency component \(X^{Lf}_{out}\) to obtain the finally result \(X_{OUT}\). This process can be formulated as follows:
\begin{equation}
\begin{aligned}
X_{OUT} = Concat(X_{\text{res}}+X^{Hf}_{out}, X^{Lf}_{out})
\end{aligned}
\end{equation}

\subsection{Loss Function}
Our network is trained in an end-to-end manner, employing a composite loss function that combines both pixel-level and perceptual quality metrics for effective supervision. The primary optimization objective utilizes the smooth L1-loss to minimize the discrepancy between the final output (O) and ground truth (G) at the pixel level. To further enhance the perceptual quality of the reconstructed images, we incorporate a feature-based perceptual loss that measures the semantic differences between the predicted output and GT in a high-dimensional feature space. This perceptual loss is computed using deep features extracted from a VGG16 network~\cite{simonyan2014very} pretrained on ImageNet, which effectively captures hierarchical visual patterns. The total loss function can be formally expressed as the weighted combination of these components:
\begin{equation}
\begin{aligned}
L_{\text{smooth L1}} = 
\begin{cases}
\frac{1}{2} E^2, & \text{if } |E| < 1 \\
|E| - \frac{1}{2} , & \text{otherwise}
\end{cases}, \\
L_{\text{perceptual}} = \mathcal{L}_{\text{MSE}}\left( \text{VGG}(O),\ \text{VGG}(G) \right), \\
L_{\text{total}} = L_{\text{smooth L1}} + \lambda L_{\text{perceptual}}.
\end{aligned}
\end{equation}
where \( E = O - G \). The weighting coefficient \(\lambda\) serves to regulate the relative contributions of the perceptual loss $L_{\text{perceptual}}$ and the $L_{\text{smooth L1}}$ within the composite objective function. Through empirical validation, we establish an optimal value of $\lambda = 0.01$, which effectively balances the trade-off between structural accuracy and perceptual quality in our final optimization objective.

% where \(\lambda\) is a weight that balances the contributions of \(L_{\text{perceptual}}\) and L1-loss on the overall loss. In our implementation, \(\lambda\) is empirically set to 0.01.

\begin{table*}[htbp]
    \centering
    \renewcommand{\arraystretch}{1.4}
    \setlength{\tabcolsep}{1.8mm}
    \caption{Comparison of different methods on Snow100K dataset. The best result and the second best result are highlighted and underlined, respectively.}
    \begin{tabular}{lcccc|cccc|cccc}
    \hline
    \toprule
    \multirow{2}{*}{Methods} & 
    \multicolumn{4}{c}{Snow100K-S} & 
    \multicolumn{4}{c}{Snow100K-M} & 
    \multicolumn{4}{c}{Snow100K-L} \\
    % \cmidrule(lr){2-5} \cmidrule(lr){6-9} \cmidrule(lr){10-13}
    & PSNR$\uparrow$ & LPIPS$\downarrow$ & SSIM$\uparrow$ & IL-NIQE$\downarrow$ 
    & PSNR$\uparrow$ & LPIPS$\downarrow$ & SSIM$\uparrow$ & IL-NIQE$\downarrow$ 
    & PSNR$\uparrow$ & LPIPS$\downarrow$ & SSIM$\uparrow$ & IL-NIQE$\downarrow$ \\
    \midrule
    DesnowNet \cite{liu2018desnownet} & 32.23 & 0.277  & 0.9500  & 20.48  & 30.68  & 0.407 & 0.9035 & 22.95 & 28.79 & 0.528 & 0.8823 & 25.34 \\
    SwinIR \cite{liang2021swinir} & 33.27 & 0.214  & 0.9521  & 19.56  & 31.44  & 0.366 & 0.9409 & 21.77 & 29.28 & 0.473 & 0.8947 & 24.12 \\
    Restormer \cite{zamir2022restormer} & 33.89 & 0.163  & 0.9534  & 18.73  & 32.42  & 0.265 & 0.9301 & 21.10 & 29.36 & 0.456 & 0.8989 & 23.48 \\
    Uformer \cite{wang2022uformer} & 31.04 & 0.335  & 0.9356  & 20.52  & 31.07  & 0.381 & 0.9134 & 22.89 & 28.76 & 0.501 & 0.8901 & 25.67 \\
    SMGRAN \cite{cheng2023snow} & 34.16 & 0.153  & 0.9610  & 18.56  & 32.47  & 0.241 & 0.9434 & 20.67 & 29.45 & 0.438 & 0.9178 & 23.45 \\
    DDMSNET \cite{zhang2021deep} & 31.46 & 0.318  & 0.9276  & 20.34  & 31.35  & 0.370 & 0.9268 & 22.61 & 28.85 & 0.498 & 0.8834 & 25.11 \\
    All-in-One \cite{Li_2020_CVPR} & 32.73 & 0.243  & 0.9512  & 19.01  & 31.82  & 0.339 & 0.9198 & 21.74 & 28.99 & 0.476 & 0.8843 & 24.12 \\
    TansWeather \cite{Valanarasu_2022_CVPR} & 32.60 & 0.247  & 0.9598  & 19.45  & 31.22  & 0.378 & 0.9073 & 21.88 & 29.14 & 0.460 & 0.8865 & 24.01 \\
    Chen et al. \cite{chen2022learning} & 32.86 & 0.210  & 0.9445  & 19.23  & 31.68  & 0.354 & 0.9101 & 21.55 & 29.33 & 0.443 & 0.9089 & 23.56 \\
    WSWG-Net \cite{zhu2023learning} & 33.56 & 0.236  & 0.9528  & 19.45  & 31.48  & 0.360 & 0.9289 & 21.68 & 29.06 & 0.461 & 0.8954 & 24.12 \\
    GridFormer\cite{wang2024gridformer} & 34.05 & 0.178  & 0.9550  & 19.20  & 32.11  & 0.280 & 0.9350 & \textbf{21.40} & 29.27 & 0.456 & 0.9060 & 23.89 \\
    MambaIR \cite{guo2024mambair} & \underline{34.62} & \underline{0.148}  & \textbf{0.9638}  & \underline{18.42}  & \underline{32.50}  & \underline{0.241} & \underline{0.9455} & 20.55 & \underline{29.47} & \underline{0.435} & \textbf{0.9185} & \underline{23.41} \\
    \midrule
    \textbf{FA-Mamba (ours)} & \textbf{34.86} & \textbf{0.137}  &\underline{0.9612}  & \textbf{18.37}  & \textbf{32.68}  & \textbf{0.237} & \textbf{0.9508} & \underline{20.42}  & \textbf{29.73} & \textbf{0.418} & \underline{0.9180} &  \textbf{23.01}\\  
    \bottomrule
    \end{tabular}
    \label{snow}
\end{table*}

\section{Experiments and Analysis}
\subsection{Experimental Settings}
\subsubsection{\textbf{Datasets}} To comprehensively evaluate the rain and snow removal performance of our FA-Mamba model, we conduct extensive experiments on both synthetic and real-world datasets, including RainDrop \cite{qian2018attentive} (1,100 image pairs with 861 for training and 239 for testing), Snow100K \cite{liu2018desnownet} (50,000 training and 50,000 test images divided into Snow100K-S/M/L subsets of 16,611/16,588/16,801 images representing light/medium/heavy snow conditions), RainDS \cite{Li2019DerainBenchmark} (150 labeled images including 98 test samples), and Snow100K-Real \cite{liu2018desnownet} (1,329 real snowy images). These datasets, collected from diverse scenarios such as monitored intersections, sidewalk views, and driving environments, enable a thorough assessment of our model's effectiveness in handling synthetic artifacts and real-world degradation patterns while testing its generalization capability across different environmental conditions.

\subsubsection{\textbf{Evaluation Metrics}} To rigorously evaluate the performance of our method across different experimental settings, we employ a comprehensive set of quantitative metrics tailored to the availability of ground truth data. For labeled datasets where reference images exist, we utilize full-reference quality assessment metrics including Peak Signal-to-Noise Ratio (PSNR~\cite{hore2010image}), Structural Similarity Index (SSIM~\cite{hore2010image}), Learned Perceptual Image Patch Similarity (LPIPS~\cite{zhang2018unreasonable}), and Integrated Local NIQE (IL-NIQE~\cite{zhang2015feature}), which collectively measure both pixel-level fidelity and perceptual quality. In scenarios involving unlabeled datasets where reference images are unavailable, we adopt no-reference image quality assessment approaches, specifically IL-NIQE~\cite{zhang2015feature} and Blind/Referenceless Image Spatial Quality Evaluator (BRISQUE~\cite{mittal2012no}), where superior perceptual quality corresponds to lower numerical values for both metrics, indicating closer alignment with natural image statistics. This dual evaluation framework thoroughly assesses our method's performance across different application scenarios while maintaining consistency with established practices in image restoration quality assessment.

\subsubsection{\textbf{Training Details}} Our experimental implementation utilizes the PyTorch framework running on NVIDIA GeForce RTX 3090 GPUs, with FA-Mamba trained using the Adam optimizer~\cite{kingma2014adam} configured with a conservative batch size of 2 to accommodate memory constraints while maintaining training stability. All training samples are uniformly resized to 256×256 to random patch cropping during the training phase, deliberately avoiding additional augmentation techniques to isolate the inherent capabilities of the model. The optimization process employs an initial learning rate of \(3 \times 10^{-4}\) for the first 100,000 iterations, followed by a reduced rate of \(1 \times 10^{-4}\) for an additional 50,000 iterations to facilitate precise convergence. For comparative analysis, we primarily leverage officially released pre-trained models or published benchmark results where available; in cases where such resources are inaccessible, we rigorously retrain baseline models on identical datasets under matched experimental conditions to ensure equitable performance comparisons. This methodology ensures both the reproducibility of our results and the validity of comparative evaluations.

\subsection{Comparisons with SOTA Methods}
To validate the efficacy of our proposed model, we conduct comprehensive experiments on two particularly challenging yet common weather-related image restoration tasks: snow removal and raindrop removal. Meanwhile, we also conduct extensive experiments on synthetic and real datasets to fully explore the effectiveness and generalization of the proposed FA-Mamba.

\begin{table}[t]
    \centering
    \renewcommand{\arraystretch}{1.4}
    \setlength{\tabcolsep}{12pt}
    \caption{Comparison of different methods on Raindrop dataset. The best result and the second best result are highlighted and underlined, respectively.}
    \resizebox{\linewidth}{!}{
    \begin{tabular}{l|c|c|c}
    \hline
        \toprule
        \textbf{Method} & \textbf{Venue} & \textbf{PSNR }$\uparrow$ & \textbf{SSIM }$\uparrow$ \\

    \hline
        
        AttnGAN \cite{Qian_2018_CVPR} & CVPR'18 & 31.59 & 0.917  \\
        DuRN \cite{Liu_2019_CVPR} & CVPR'19 & 31.24 & 0.926  \\
        RainAttn \cite{Quan_2019_ICCV} & ICCV'19 & 31.44 & 0.926    \\
        All-in-One \cite{Li_2020_CVPR} & CVPR'20   & 31.12 & 0.930    \\
        CCN \cite{Quan_2021_CVPR} & CVPR'21 & 31.44 & \underline{0.947}  \\
        IDT \cite{9798773} & TPAMI'22 & 31.87 & 0.931   \\
        TransWeather \cite{Valanarasu_2022_CVPR} & CVPR'22 & 30.96 & 0.923  \\
        WeatherDiff \cite{10021824} & TPAMI'23 & 32.13 & 0.939   \\
        DRSformer \cite{Chen_2023_CVPR} & CVPR'23 & 32.40 & 0.940 \\
        PatchDM \cite{10021824} & TPAMI'23 & 32.31 & 0.946\\
        WaveDM \cite{10420512} & TMM'24 & 32.25 & \textbf{0.948}  \\
        GridFormer \cite{wang2024gridformer} & IJCV'24 &   \underline{32.92}& 0.940    \\
        MambaIR \cite{guo2024mambair} & ECCV'24 &   32.15& 0.938  \\
        %& VMambaIR [36] & arxiv'24 &   &   &   &   &    \\
        \midrule
        \textbf{FA-Mamba(ours)} & - & \textbf{33.18} & 0.941   \\
        \bottomrule
    \hline
    \end{tabular}
    }
    \label{Raindrop}
\end{table}

\subsubsection{\textbf{Synthetic Datasets}} The quantitative evaluation on the Snow100K dataset presented in TABLE~\ref{snow}, reveal that our method achieves superior image quality with remarkable fidelity to ground truth across all test subsets. FA-Mamba establishes new benchmarks with PSNR/LPIPS scores of 34.86/0.137, 32.68/0.237, and 29.73/0.418 for Snow100K-S, Snow100K-M, and Snow100K-L respectively. Similarly, the experimental results on the Raindrop dataset, shown in TABLE~\ref{Raindrop}, demonstrates that FA-Mamba surpasses the current state-of-the-art method GridFormer~\cite{wang2024gridformer} by a significant margin of 0.26dB. 

\begin{figure*}[htbp]
\centering
\includegraphics[width=1\textwidth]{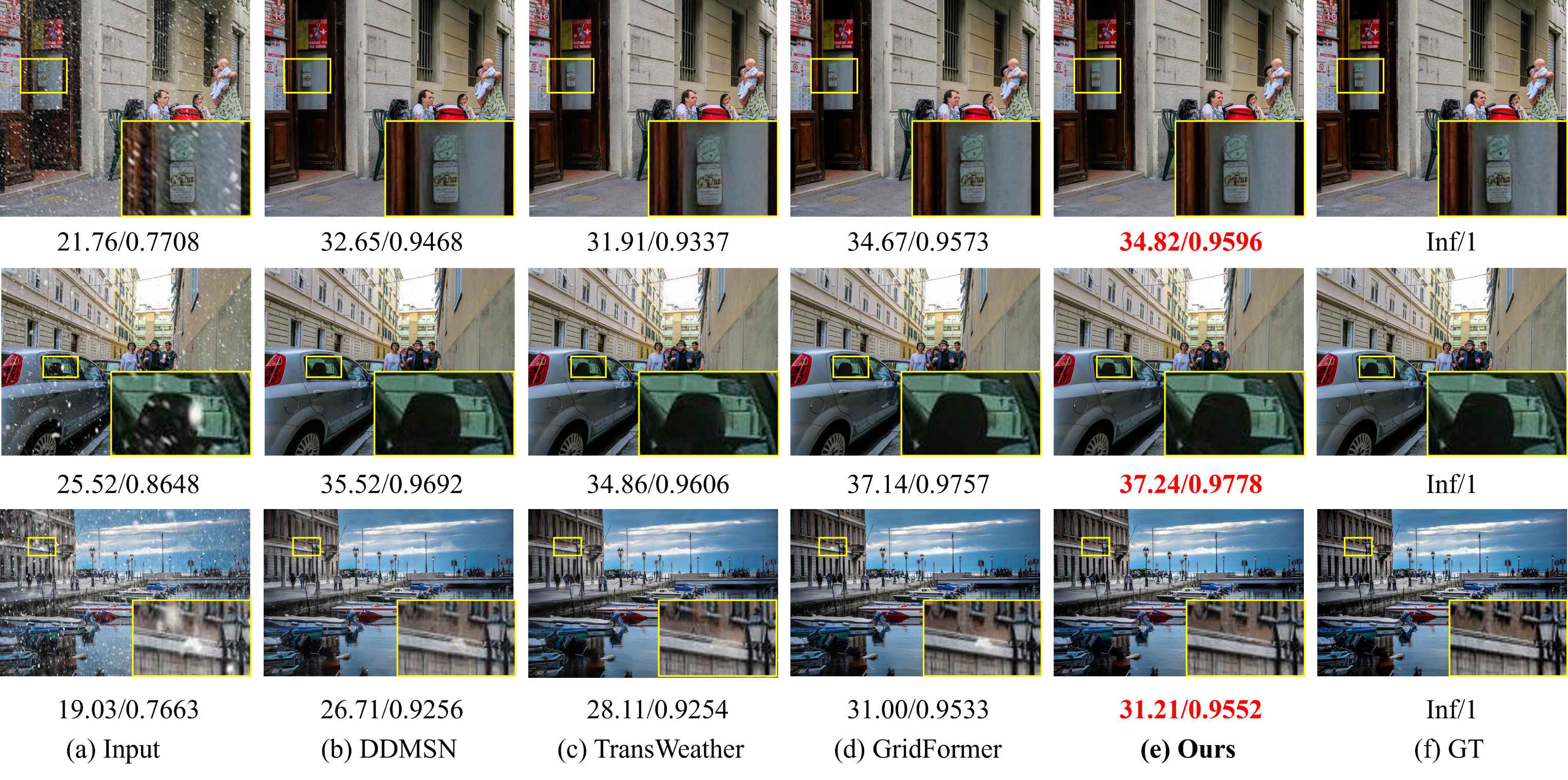}
\caption{Image desnowing comparison on Snow100K-S, Snow100K-M, and Snow100K-L test sets (from top to bottom). (a) snow images; (b)-(d) images restored by DDMSN, TransWeather, and GridFormer, respectively; (e) images restored by our proposed FA-Mamba; (f) clean images. Please zoom in for the best view.}
\label{fig:visual comparison on image snow removal}
\end{figure*}

\begin{figure*}[htbp]
\centering
\includegraphics[width=1\textwidth]{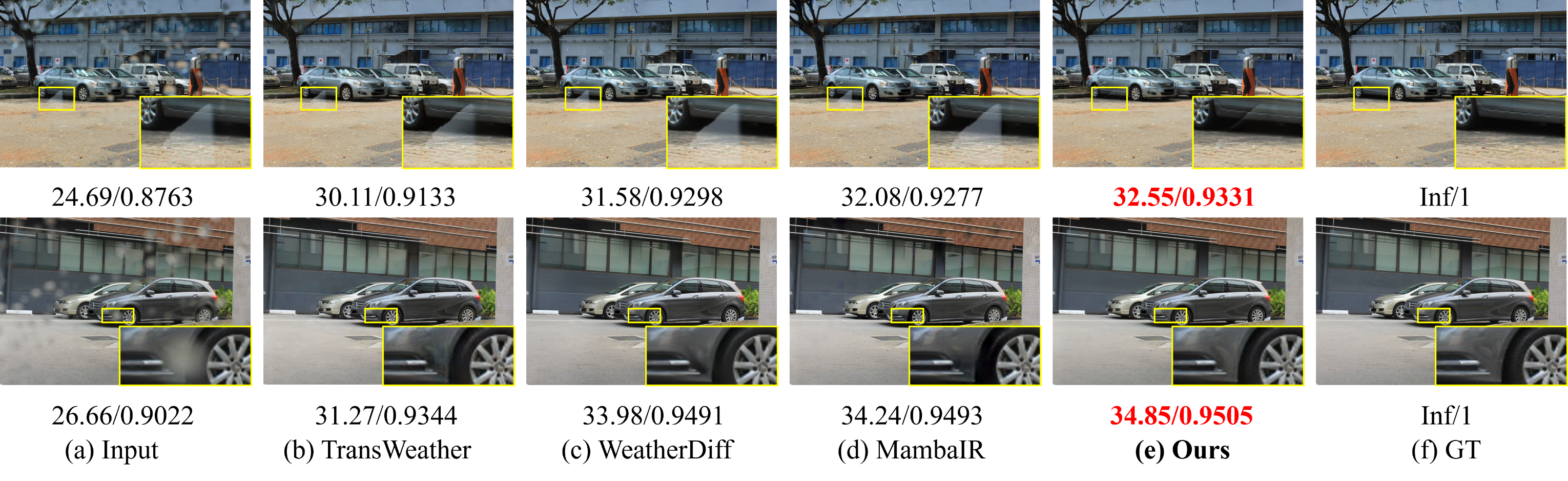}
\caption{Visual comparison on raindrop dataset. (a) raindrop images; (b)-(d) images restored by TransWeather, WeatherDiff, and MambaIR, respectively; (e) images restored by our proposed FA-Mamba; (f) clean images. Please zoom in for the best view.}
\label{fig:visual comparison on image raindrop removal.}
\end{figure*}

Qualitative comparisons, illustrated in Figs.~\ref{fig:visual comparison on image snow removal} and ~\ref{fig:visual comparison on image raindrop removal.}, highlight that existing methods frequently exhibit incomplete removal of precipitation artifacts and introduce undesirable distortions in heavily occluded regions. In contrast, FA-Mamba not only achieves comprehensive precipitation removal but also faithfully reconstructs intricate background textures. This performance advantage stems from two key innovations: (1) the high-frequency guidance mechanism implemented in the PGB module, and (2) the DFEB's exceptional capability in modeling regional feature correlations. \textbf{Notably, FA-Mamba secured third place in LPIPS evaluation in the NTIRE 2025 Challenge on day and night raindrop removal for dual-focused images~\cite{Li2025NTIRE2C}, further validating its effectiveness in real-world scenarios.}

\begin{table}[http]
\renewcommand{\arraystretch}{1.3}
\caption{Quantitative comparison of IL-NIQE and BRISQUE scores on a real-world image dataset. The best result and the second best result are highlighted and underlined, respectively.}
\centering
\renewcommand{\arraystretch}{1.4}
\resizebox{\linewidth}{!}{
\begin{tabular}{lcccc}
\toprule
& \multicolumn{2}{c}{\textbf{\underline{Snow100K-Real\cite{liu2018desnownet}}}} & \multicolumn{2}{c}{\textbf{\underline{RainDS\cite{quan2021removing}}}} \\ 
\textbf{} & IL-NIQE $\downarrow$ & BRISQUE $\downarrow$ & IL-NIQE $\downarrow$ & BRISQUE $\downarrow$ \\ \hline
TransWeather\cite{Valanarasu_2022_CVPR} & \underline{22.12} & \textbf{15.52} & 22.32 & 22.78 \\ 
Gridformer\cite{wang2024gridformer} & 22.06 & 19.54 & \underline{22.14} & \underline{19.97} \\ \hline
\textbf{FA-Mamba} & \textbf{21.88} & \underline{17.75} & \textbf{21.79} & \textbf{15.79} \\ 
\bottomrule
\end{tabular}
}
\label{real}
\end{table}

% \begin{table}[htbp]
% % \setlength{\tabcolsep}{4pt} % 默认是 6pt
% \renewcommand{\arraystretch}{1.5}  % 增加行高
% \caption{\textbf{NIQE and BRISQUE Results Under Real-World Scenarios.}}
% \begin{center}
% \resizebox{\linewidth}{!}{
% \begin{tabular}{c|cccc|c}
% \toprule
% Methods & Raindrop/Snow  & Transweather\cite{Valanarasu_2022_CVPR} & MambaIR \cite{guo2024mambair} & Gridformer\cite{wang2024gridformer} & FA-Mamba \\
% \hline
% IL-NIQE $\downarrow$  & 27.51/21.92 & 27.41/21.99 &  &  29.29/22.06   & 31.23/21.88 \\
% \hline
% BRISQUE $\downarrow$ &30.97/19.63 & 20.05/15.52 &  &  21.83/19.54  & 7.66/17.75\\
% \bottomrule
% \end{tabular}
% }
% \label{tab:ablation}
% \end{center}
% \end{table}

\begin{figure*}[ht]
\centering
\includegraphics[width=1\textwidth]{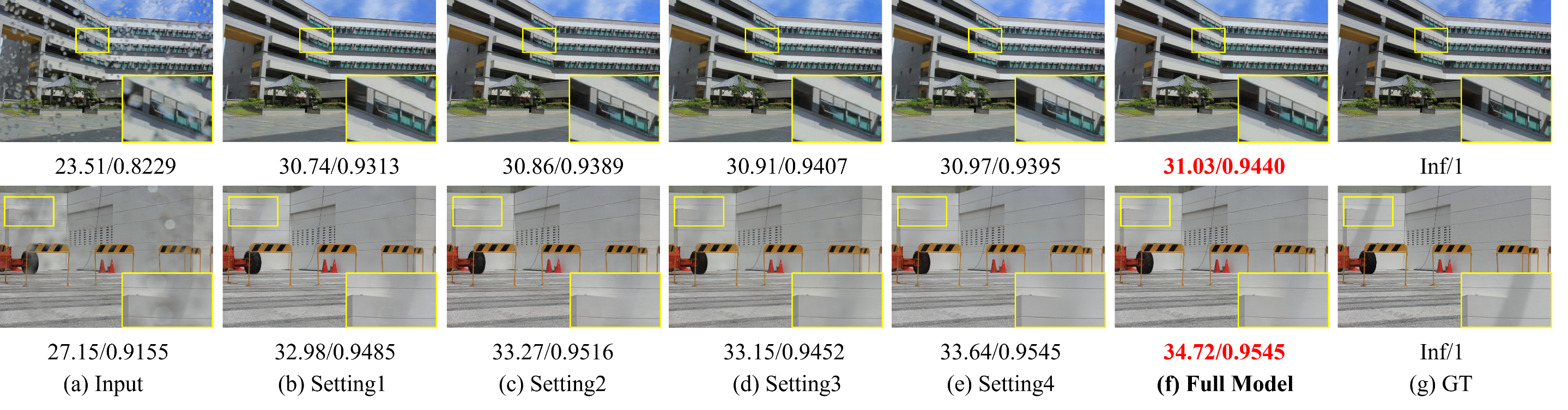}
\caption{Visual comparison of image raindrop removal in ablation studies. (a) raindrop images; (b)-(d) restored images by Setting1, Setting2, Setting3, and Setting4, respectively; (f) restored images by our full model (FA-Mamba); (f) clean images.}
\label{fig:Visualization of Ablation Study Results.}
\end{figure*}

\begin{figure}[ht]
\centering
\includegraphics[width=\linewidth]{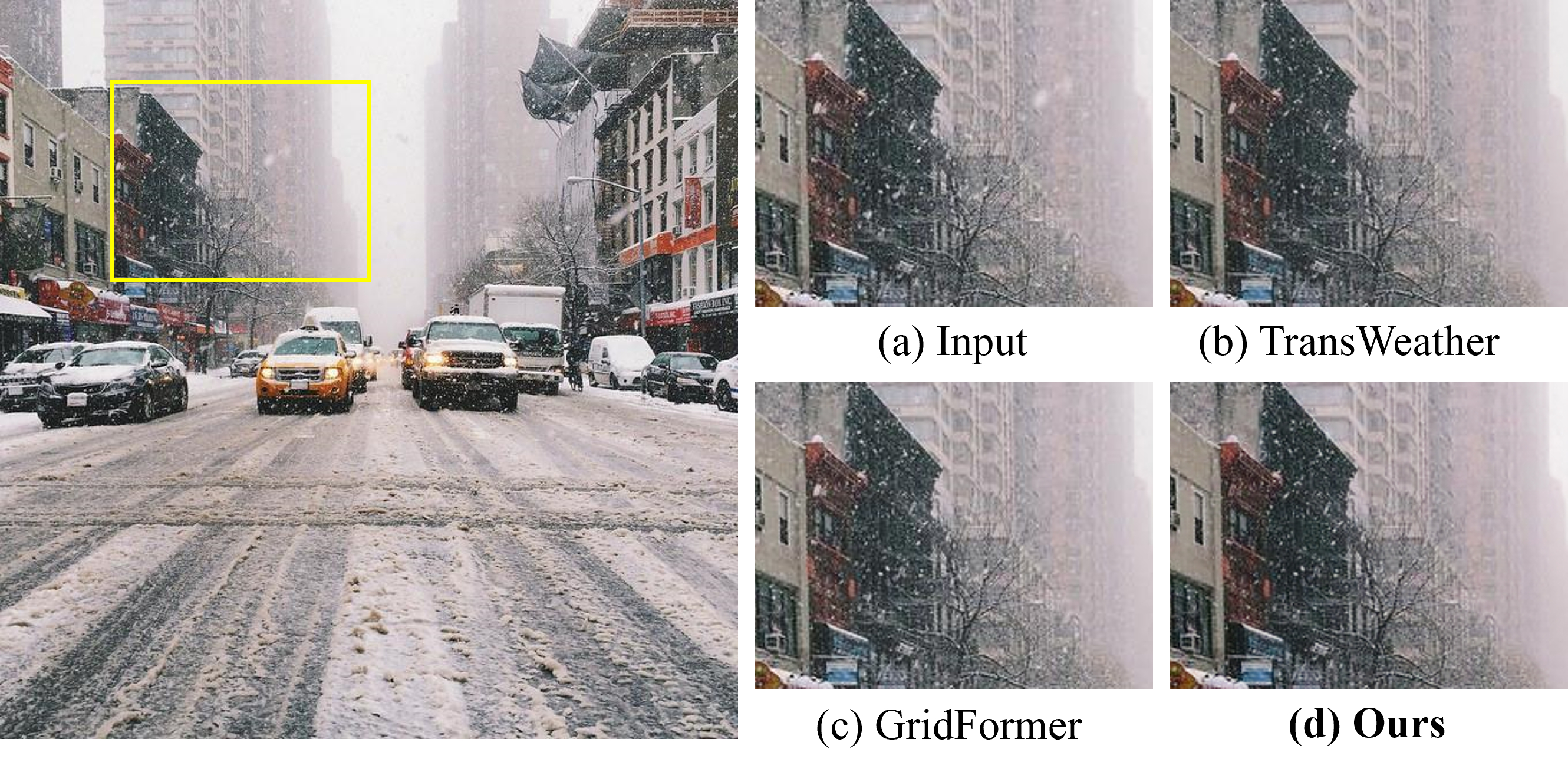}
\caption{Desnowing comparison on Snow100K-Real. (a) snow image; (b)-(d) restored images by TransWeather, GridFormer, and our FA-Mamba, respectively.}
\label{fig:visual comparison on image snow-real removal}
\end{figure}

\begin{figure}[ht]
\centering
\includegraphics[width=\linewidth]{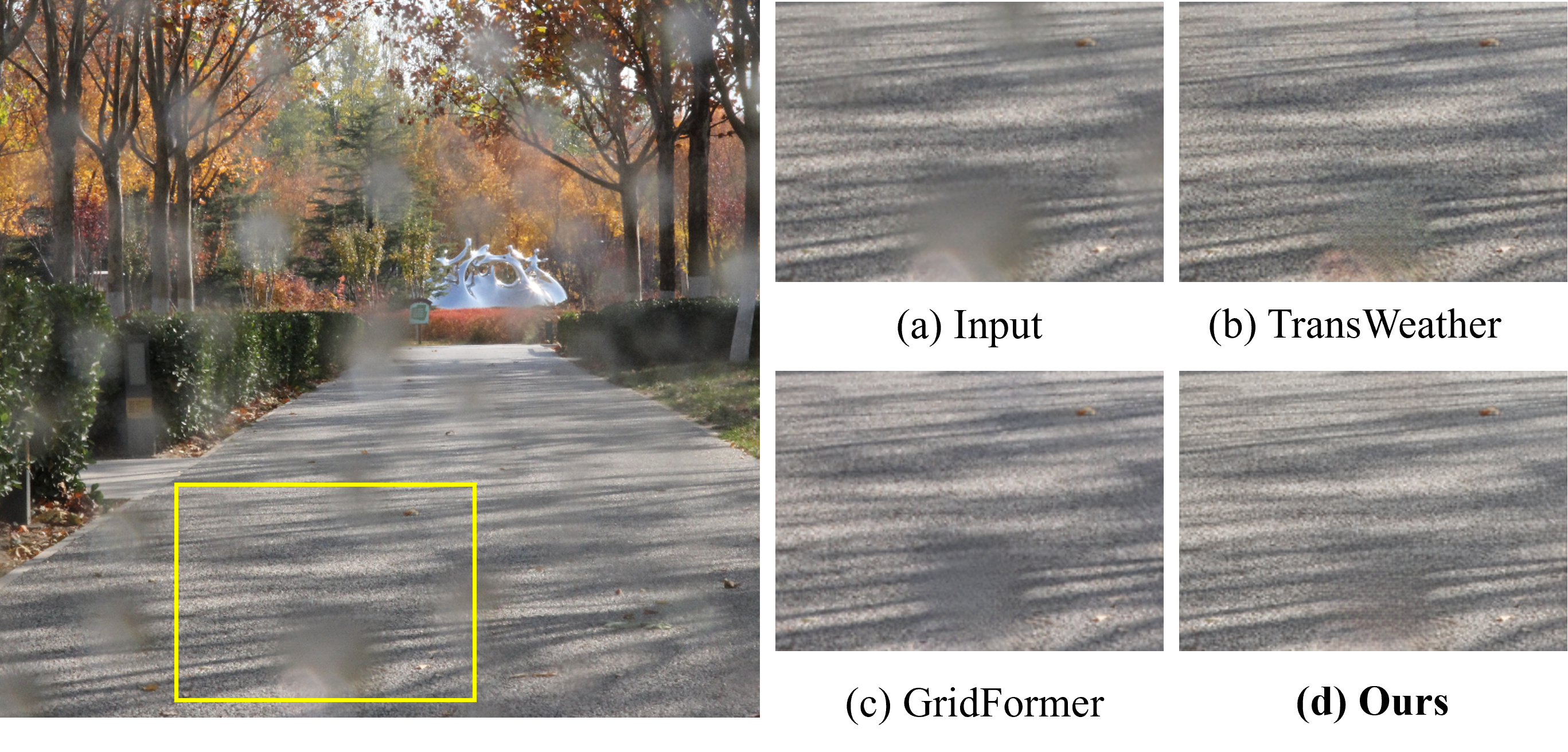}
\caption{Deraining comparison on RainDS. (a) raindrop image; (b)-(d) restored images by TransWeather, GridFormer, and our FA-Mamba, respectively.}
\label{fig:visual comparison on image raindrop-real removal}
\end{figure}

\subsubsection{\textbf{Real-world Datasets}} To further verify the generalization ability of the proposed method, we tested its performance on real images (Snow100K-Real\cite{liu2018desnownet} and RainDS~\cite{Li2019DerainBenchmark} datasets) and compared it with state-of-the-art weather restoration networks, including the classical TransWeather~\cite{Valanarasu_2022_CVPR} and the latest weather restoration network GridFormer~\cite{wang2024gridformer}. As shown in TABLE~\ref{real}, we provide a quantitative comparison across these two datasets using reference-free image quality metrics (IL-NIQE and BRISQUE). The experimental results demonstrate that our model achieves superior perceptual quality, exhibiting statistically significant improvements over the current state-of-the-art weather restoration model, GridFormer, on all evaluated datasets. Meanwhile, we present comprehensive visual comparisons of different methods on real-world scenarios. Fig.~\ref{fig:visual comparison on image snow-real removal} showcases desnowing performance on challenging cases from the Snow100K-Real~\cite{liu2018desnownet} dataset, where our model demonstrates superior capability in preserving fine details while effectively removing snow artifacts. Similarly, Fig.~\ref{fig:visual comparison on image raindrop-real removal} illustrates raindrop removal results, particularly highlighting our model's robustness in handling severe cases with intense raindrop artifacts that remain problematic for both TransWeather and GridFormer. The visual comparisons clearly demonstrate the advantages of our proposed model in terms of artifact removal completeness and detail preservation.

\begin{table}[t]
\renewcommand{\arraystretch}{1.4}
\caption{\textbf{Ablation studies of different components.}}
\begin{center}
\resizebox{\linewidth}{!}{
\begin{tabular}{c|ccccc|c|c}
\toprule
Experiment & DFEB & HFEM & PGB & 2D-Scan & AF-Scan & PSNR & SSIM \\ \hline
Baseline  &  &  &  &  &  & 31.44 & 0.928\\
Setting1  & \checkmark & & & \checkmark &  &  32.30 & 0.934 \\
% Setting3  & \checkmark & \checkmark & & 32.78 & 0.938 \\
Setting2  & \checkmark &\checkmark &  & \checkmark& &32.62  & 0.936 \\
Setting3  & \checkmark &  & \checkmark & \checkmark& & 32.59 & 0.937 \\
% Setting2  & \checkmark & & \checkmark & & & 32.78 & 0.938 \\
Setting4  & \checkmark & \checkmark&\checkmark & \checkmark & & 33.04 & 0.938 \\
\hline
Full Model  & \checkmark &\checkmark &\checkmark & & \checkmark & 33.18 & 0.941 \\
\bottomrule
\end{tabular}
}
\label{Ablation}
\end{center}
\end{table}

\begin{table}[t]
\renewcommand{\arraystretch}{1.4}
\caption{\textbf{Study of the combination strategy of DFEB on the raindrop dataset.}}
\begin{center}
\resizebox{\linewidth}{!}{
\begin{tabular}{cccccccc}
\toprule
Case &Branch1  & Branch2 & Params & FLOPs & PSNR & SSIM \\
\hline
1  & Conv & Trans & 11.35M & 190G & 32.11 & 0.932\\
2  & Conv & Mamba & 11.47M & 189G & 32.30 & 0.934\\
\bottomrule
\end{tabular}
}
\label{combination strategy}
\end{center}
\end{table}

\subsection{Ablation Study}
To ensure consistent and reliable performance evaluation, we conduct our ablation studies on the Raindrop dataset using a standardized experimental configuration with network depth parameters of \{6,6,4,4\} and channel dimension $C$ is set to 180. This controlled setup enables fair and systematic comparison of different model components.

\subsubsection{\textbf{Ablation on DFEB}} The Dual-Branch Feature Extraction Block (DFEB) serves as a fundamental component of FA-Mamba, specifically designed for robust feature extraction in image reconstruction. To evaluate its effectiveness, we perform comprehensive ablation studies. As demonstrated in TABLE~\ref{Ablation}, Setting 1 incorporating only DFEB modules, which achieve a 0.86 dB PSNR improvement over the Baseline consisting of conventional CNN branches. Further investigation into various integration strategies reveals that, as shown in TABLE~\ref{combination strategy}, Case 2 provides a 0.19 dB PSNR enhancement compared to Case 1 while maintaining similar model complexity and computational requirements. These experimental results conclusively validate both the necessity and efficacy of the CNN-Mamba integration strategy in our framework.

% \begin{table}[htbp]
% \renewcommand{\arraystretch}{1.5}
% \caption{\textbf{Study of the combination strategy of CNN and Mamba on raindrop dataset.}}
% \begin{center}
% \resizebox{\linewidth}{!}{
% \begin{tabular}{cccccccc}
% \toprule
% Case & Branch1  & Branch2 & Params & GPU & FLOPs & PSNR & SSIM \\
% \hline
% 1  & Conv & Conv & 5.64 M &  2066 MiB & 136 G &  & \\
% 2  & Conv & Trans & 11.35 M &  2132 MiB    & 190 G & 32.11 &0.932\\
% 3  & Conv & Mamba &  11.47 M & 3700 MiB  &  189 G  & 32.30 & 0.934\\
% \bottomrule
% \end{tabular}
% }
% \label{tab:ablation}
% \end{center}
% \end{table}

\begin{figure*}[htbp]
\centering
\includegraphics[width=1\textwidth]{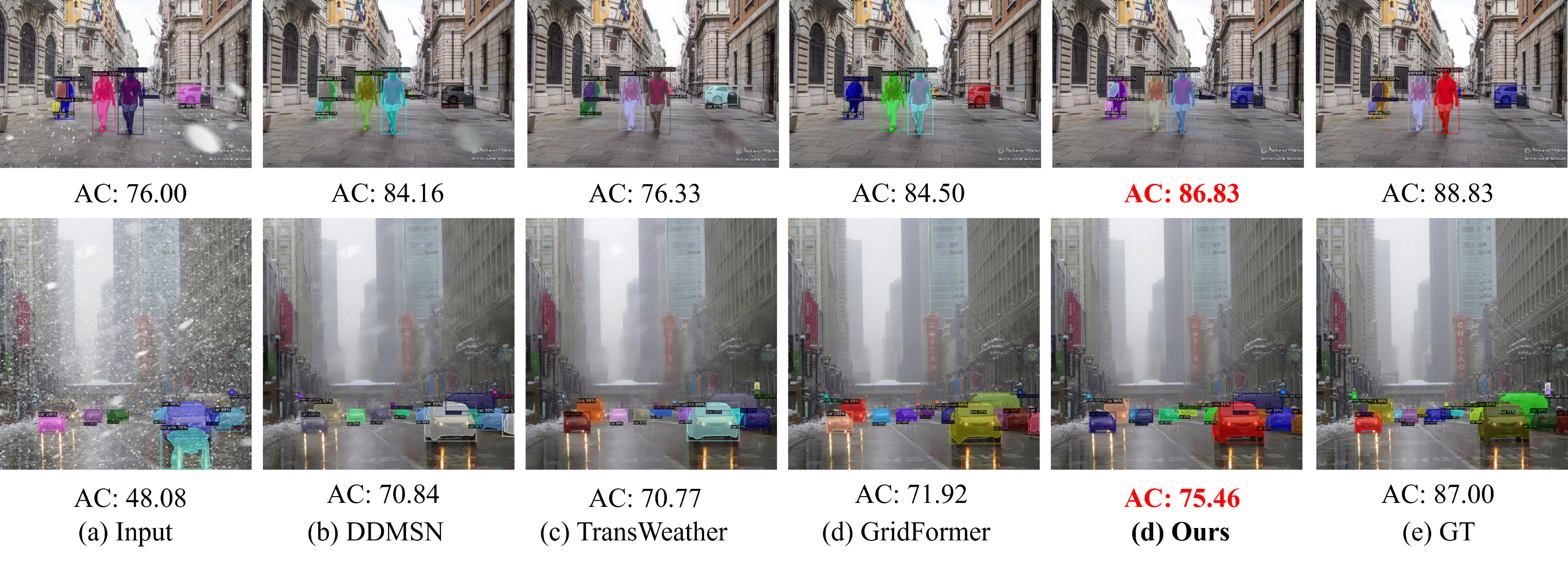}
\caption{Detection performance evaluated using the Detectron2 framework. AC represents the Average Confidence score. (a) recognition result on the snow image; (b)-(d) recognition result on the restored image of DDMSN, TransWeather, and GridFormer, respectively; (e) recognition result on the restored image of our FA-Mamba; (f) recognition result on the clean image.}
\label{fig:recognition result on the
snow image}
\end{figure*}

\begin{figure*}[htbp]
\centering
\includegraphics[width=1\textwidth]{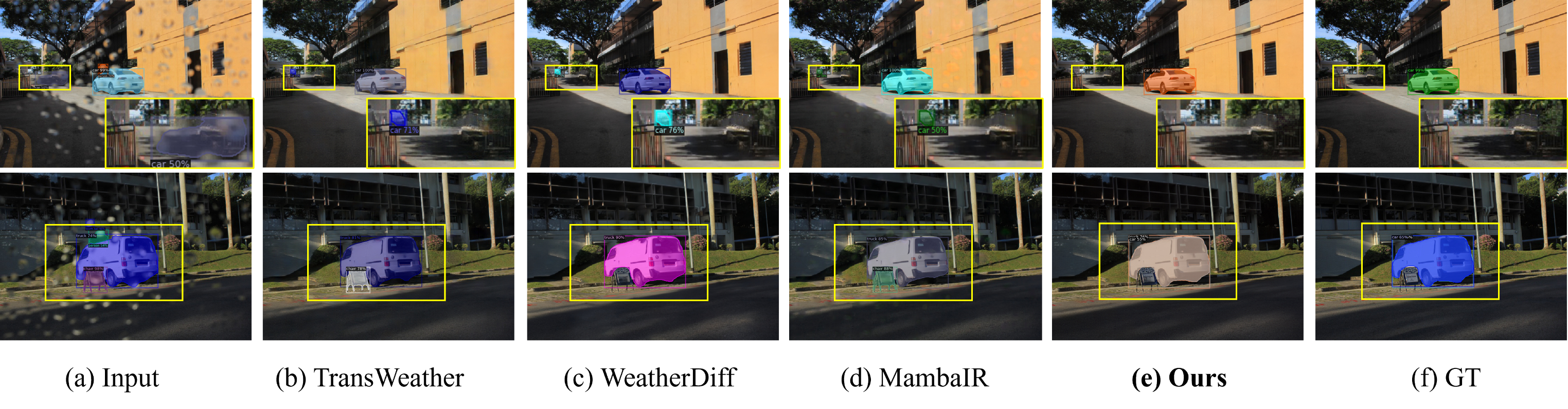}
\caption{Detection performance evaluated using the Detectron2 framework. (a) recognition result on the raindrop image; (b)-(d) recognition
result on the restored image of TransWeather, WeatherDiff, and MambaIR, respectively; (e) recognition result on the deraining image of our FA-Mamba; (f) is the recognition result on the clean image.}
\label{fig:recognition result on the
raindrop image}
\end{figure*}

\begin{figure}[htbp]
\centering
\includegraphics[width=\linewidth]{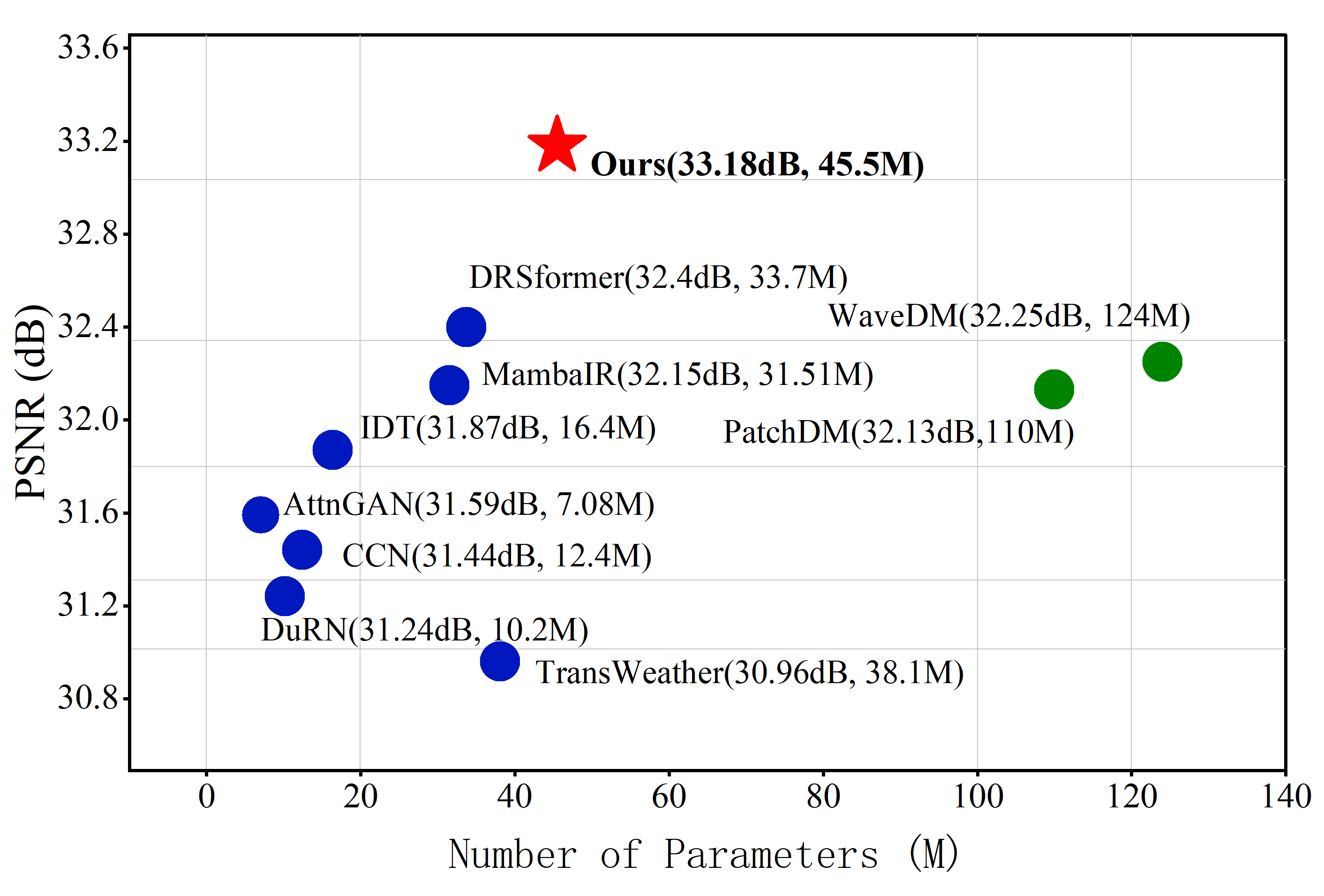}
\caption{Model parameters and performance comparison based on the raindrop dataset}
\label{size}
\end{figure}

\subsubsection{\textbf{Ablation on PGB and HFEM}} Our ablation study systematically investigates the synergistic effects of the Prior-Guided Block (PGB) and High-Frequency Enhancement Module (HFEM), as presented in TABLE~\ref{Ablation}. The experimental results reveal that Setting 2, which integrates HFEM to concatenate extracted high-frequency priors with DFEB's high-frequency features, yields a 0.32 dB PSNR improvement over Setting 1. Similarly, Setting 3 demonstrates that incorporating PGB alone provides an equivalent 0.29 dB enhancement. Notably, Setting 4 achieves optimal performance by combining both modules, with comprehensive analysis showing that HFEM contributes refined high-frequency texture information while PGB effectively utilizes this information to guide DFEB's feature extraction process. These findings quantitatively validate the complementary nature and individual importance of both components in our architecture.

\subsubsection{\textbf{Ablation on AFSM}} The Adaptive Frequency Scanning Mechanism (AFSM) is specially designed for the wavelet-based Mamba architecture, which facilitates frequency domain scanning across different subgraphs to fully exploit the inherent texture distribution characteristics in the sub-graph structure. Quantitative results in TABLE~\ref{Ablation} demonstrate that AFSM contributes a significant 0.14 dB PSNR improvement, which attributed to the ability of the mechanism to establish comprehensive frequency aware processing across all frequency blocks through its innovative AF scanning operation. This enhancement confirms the effectiveness of our frequency adaptive method in image restoration tasks.

Through the zoomed boxes in Fig.~\ref{fig:Visualization of Ablation Study Results.}, the recovered results of the full model tend to be clearer. Overall, our model outperforms all other configurations, demonstrating that each design strategy contributes to the final performance of FA-Mamba.

\subsection{Model Complexity Analysis}
The experimental results demonstrate that our model consistently outperforms most competing methods in both quantitative metrics and qualitative assessments. Furthermore, to comprehensively evaluate model efficiency, we analyze the trade-off between computational complexity and restoration performance. As illustrated in Fig.~\ref{size}, which compares parameter counts against performance metrics, our FA-Mamba achieves best result while maintaining comparable model complexity. This indicates that our FA-Mamba establishes an optimal balance between model compactness and restoration quality, making it particularly suitable for practical applications where both performance and efficiency are crucial considerations.

\begin{figure}[t]
\centering
\includegraphics[width=\linewidth]{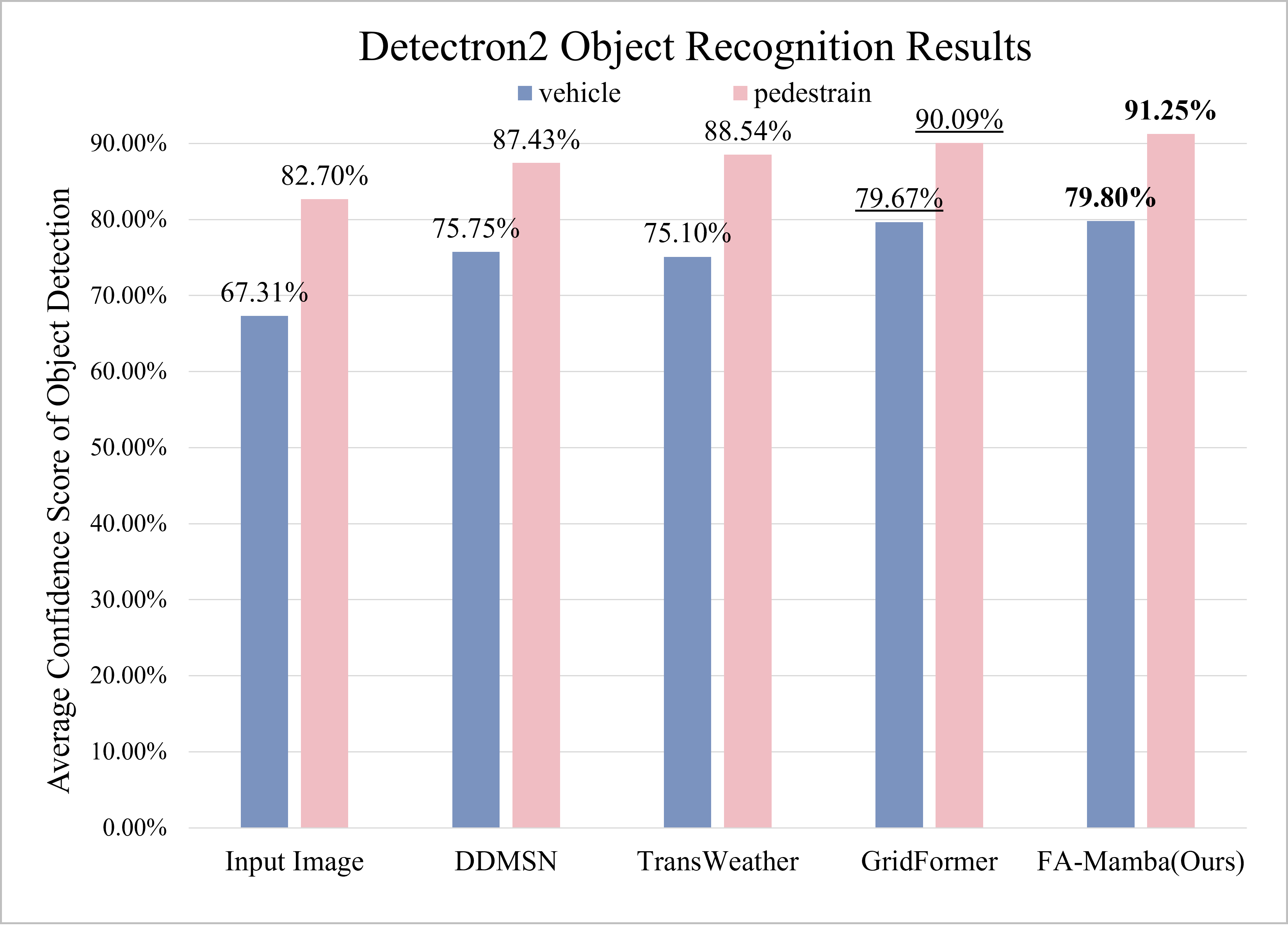}
\caption{We evaluate the average recognition confidence for vehicles and pedestrians using Google Colaboratory (Colab). We test 50 sets of traffic and pedestrian images in snowy days respectively, as well as desnowing images of ours and three deep learning-based methods. Note that the confidence of mis-recognized or unecognized target is set to 0.}
\label{fig:Average Confidence Score of Object Detection}
\end{figure}

\begin{figure}[t]
\centering
\includegraphics[width=\linewidth]{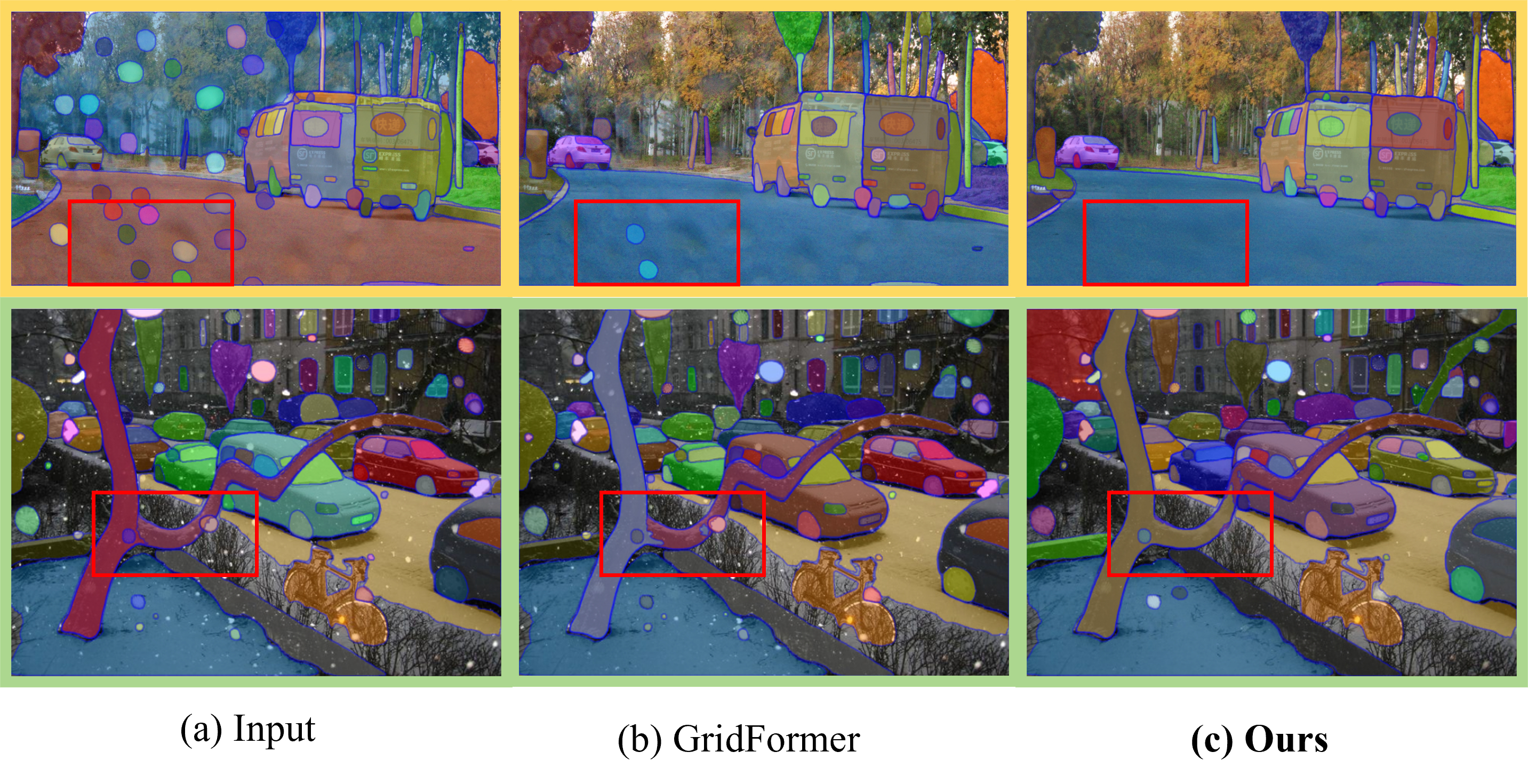}
\caption{Segmentation performance evaluated using the SAM framework. The first row shows the segmentation results on raindrop images, while the second row shows the segmentation results on snow images.}
\label{fig:SAM anything}
\end{figure}

\subsection{Traffic Application}
A critical challenge in Intelligent Transportation Systems involves the accurate recognition and segmentation of vehicles and pedestrians in weather-degraded images. To validate the practical utility of our FA-Mamba, we conduct comprehensive evaluations using two state-of-the-art vision models. First, we employ Detectron2~\cite{wu2019detectron2}, pre-trained on the COCO dataset, to quantitatively assess the impact of weather removal on object detection performance. Second, we leverage the Segment Anything Model (SAM~\cite{kirillov2023segment}) to evaluate the segmentation results of the restored images.

Specifically, we visualize desnowed and derained images along with their corresponding recognition results. Our evaluation includes the inputs, ground truths, and restored images from various weather removal methods, allowing for a comprehensive assessment of performance across different weather conditions, as demonstrated in Figs.~\ref{fig:recognition result on the snow image} and ~\ref{fig:recognition result on the raindrop image}. It can be seen that our FA-Mamba can effectively remove the weather degradation effects encountered in traffic images while preserving important object information that plays a crucial role in recognition by the Google Colaboratory API. Meanwhile, we test the Detectron2 object detection model on a set of 50 images we collected containing traffic and pedestrians in rainy and snowy conditions. We evaluate and record the object recognition results of vehicles and pedestrians, which have the highest  average confidence scores in their respective classification on Google Colaboratory API, respectively, labeled as vehicle and pedestrian. The comparison results of object recognition is shown in Figs.~\ref{fig:Average Confidence Score of Object Detection}. It can be seen that our FA-Mamba improves the average confidence of Google Colaboratory API for vehicle and pedestrian recognition on degraded images to 79.80\% and 91.25\%. 

Furthermore, we also used SAM~\cite{kirillov2023segment} to segment the restored images under different weather conditions. As illustrated in Fig.~\ref{fig:SAM anything}, the segmentation results on our restored images exhibit significant improvements in both boundary precision and region consistency compared to the degraded inputs, quantitatively demonstrating that the restoration capability of FA-Mamba directly enhances downstream segmentation performance by effectively preserving critical structural information while removing weather artifacts.

\section{Conclusion}
In this work, we proposed Frequency-Aware Mamba (FA-Mamba), an innovative framework that synergistically combines frequency-domain guidance with advanced sequence modeling for high-performance image restoration. The proposed architecture comprises two fundamental components: (1) a Dual-Branch Feature Extraction Block (DFEB) that establishes comprehensive local-global interactions through bidirectional 2D frequency-adaptive scanning; and (2) a Prior-Guided Block (PGB) that employs wavelet-based high-frequency residual learning to precisely reconstruct texture details. Furthermore, we introduce a groundbreaking Adaptive Frequency Scanning Mechanism (AFSM) that extends the Mamba architecture's capabilities to perform frequency-domain scanning across multiple subgraphs, effectively exploiting the inherent texture distribution patterns within each sub-band structure. Comprehensive experimental evaluations validate both the computational efficiency and restoration effectiveness of our FA-Mamba framework.

\bibliographystyle{unsrt}  
%\bibliography{references} 

\end{document}